%% file: arxiv_preprint.tex
\documentclass[11pt]{article}
\usepackage[preprint]{acl}
\usepackage{times}
\usepackage{latexsym}
\usepackage[T1]{fontenc}
\usepackage[utf8]{inputenc}
\usepackage{microtype}
\usepackage{inconsolata}
\usepackage{graphicx}
\usepackage{float}
\usepackage{stfloats}
\usepackage{booktabs}
\usepackage{array}
\usepackage{tabularx}
\usepackage{makecell}
\usepackage{amsmath}
\usepackage{amssymb}
\usepackage{xspace}
\usepackage{url}
\usepackage{xcolor}
\usepackage{listings}
\usepackage{enumitem}
\usepackage{subcaption}
\usepackage{placeins}
\usepackage{flushend}

\title{Eval-Pair Matrix: Answer-Paired Meta-Evaluation of LLM Judges for Grounded RAG}
\author{Sriram Selvam\\\texttt{selvamsriram@gmail.com}\And Anneswa Ghosh\\\texttt{anneswaghosh@gmail.com}}
\newcommand{\method}{Eval-Pair Matrix\xspace}

\newcolumntype{Y}{>{\raggedright\arraybackslash}X}
\newcolumntype{L}[1]{>{\raggedright\arraybackslash}p{#1}}
\begin{document}
\maketitle
\input{main_body}

\input{references}
\input{appendix_content}
\end{document}

%% file: main_body.tex
\begin{abstract}
LLM-as-a-judge evaluation is widely used for retrieval-augmented generation (RAG), but reusing the same model family as both generator and judge makes self-leniency difficult to identify. We introduce \method, a controlled meta-evaluation protocol for source-grounded RAG. Starting from GaRAGe questions and grounding passages, we induce one hidden answer-causal contradiction per record, generate answers from perturbed passages with GPT, Grok, and Gemini models, and then use the same models as blind judges to evaluate each answer against the original passages. The experiment contains 300 core records, 897 labeled generator outputs, and 2,683 judge verdicts in a crossed $3\times3$ matrix; the primary analysis uses 275 fully validated records. Instead of comparing diagonal and off-diagonal cells across different answers, we estimate same-model effects by pairing judges on the exact same candidate answer. This changes the interpretation: diagonal and off-diagonal F1 are similar, and the paired same-model recall effect is near zero ($-0.5$ pp; 95\% cluster-bootstrap CI $[-2.7,+1.7]$). The only robust paired gap is lower matching-judge flagging for answers that avoided the induced claim ($-4.3$ pp). A targeted human evaluation finds that reviewed apparent false positives are alternate source-error detections, mistakes in labeling whether the induced claim was adopted, or unclear cases; none were adjudicated as genuine false alarms. The lesson is methodological: RAG judge studies should report full matrices, answer-paired effects, behavior strata, and label-task alignment.
\end{abstract}

\section{Introduction}

Retrieval-augmented generation (RAG) systems combine a language model's parametric knowledge with external passages supplied at inference time \citep{lewis2020rag}. This makes them attractive for knowledge-intensive applications, but it also creates a difficult evaluation problem: an answer may be fluent, cited, and still contradict the reference passages. In practice, many RAG evaluations now use another LLM as the judge because LLM judges are cheap, fast, and can emit structured rationales \citep{liu2023geval,zheng2023judging,kim2023prometheus}. When the generator and judge are from the same model family, evaluation may become circular.

Existing LLM-as-a-judge work documents position, verbosity, length, style, familiarity, and self-preference effects \citep{dubois2024length,shi2024position,liu2024narcissistic,wataoka2024selfpreference,ye2024justice,spiliopoulou2025playfavorites}. This literature also shows that self-preference is an identification problem: a judge may favor its own model's output because it is better, clearer, more familiar, or more stylistically aligned, rather than because the judge is lenient toward its own model. We ask how this problem changes in source-grounded RAG. If we induce a known source-relative contradiction hidden from the judge, then the judge is not choosing a favorite answer; it is checking whether a candidate answer contradicts the original evidence.

\begin{figure*}[t]
  \centering
  \includegraphics[width=0.96\linewidth]{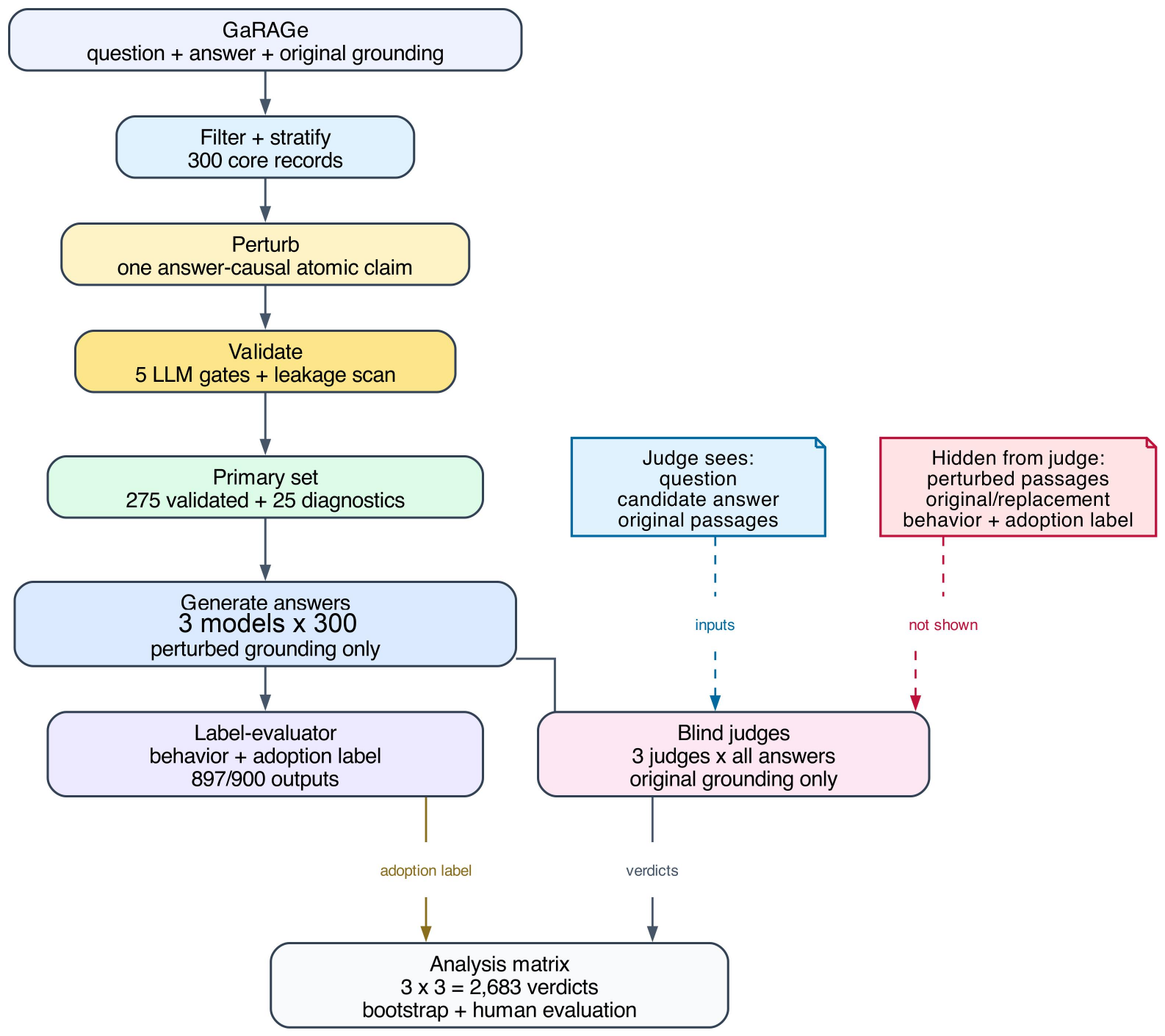}
  \caption{The \method pipeline. Perturbations are generated and validated before answer generation. Judges see only the original question, candidate answer, and original grounding; they do not see the perturbation, behavior label, or induced-error adoption label. The final analysis combines the crossed matrix, answer-paired bootstrap, behavior stratification, and a targeted human evaluation.}
  \label{fig:pipeline}
\end{figure*}

We build the task from GaRAGe, a benchmark with human-written answers and passage-level grounding annotations \citep{sorodoc2025garage}. For each selected example, an LLM perturber chooses one answer-causal target proposition and rewrites every passage that states or entails that claim so the evidence consistently supports a plausible but false replacement value. Generators answer the original question using the perturbed grounding. Judges then evaluate the answer against the unmodified grounding only. Figure~\ref{fig:pipeline} summarizes the end-to-end design.

A raw diagonal analysis compares each judge's matching-model cell with that judge's two off-diagonal generator cells. That contrast is useful descriptively but not inferentially: it compares different answers with different styles, refusal rates, and difficulty. The paired estimand instead asks: \emph{on the exact same candidate answer}, does the matching judge behave differently from the non-matching judges? This reframes the study from provider-specific answer distributions to same-model pairing effects.

This paired analysis changes the interpretation of the raw matrix. On the 275 fully validated records, the raw diagonal/off-diagonal F1 difference is small (84.4 diagonal vs. 83.4 off-diagonal). A paired answer-level cluster bootstrap finds no robust same-model effect on error-detection recall ($\Delta=-0.5$ pp, 95\% CI $[-2.7,+1.7]$): judges are not broadly less likely to catch induced errors in answers generated by their own model. The only robust paired difference appears on answers that avoided the induced false claim: the judge from the same model as the generator flags these answers less often than the two other judges ($\Delta=-4.3$ pp, CI $[-6.6,-2.0]$). This should not be read as a lower false-alarm rate. Our automatic induced-error adoption label records only whether the answer adopted the false claim introduced by the perturbation, while the judge is asked to flag any factual error against the original passages. An answer can avoid the induced false claim and still contain another source-relative error, especially when it refuses, hedges, or describes inconsistent evidence. In the targeted 88-case human evaluation, none of the 25 reviewed apparent false positives were genuine false alarms: they were alternate real source errors, mistakes in the induced-error adoption label, or unclear cases. Thus the apparent false-positive-rate effect is best interpreted as a mismatch between the adoption label and the judge's broader source-grounding task, not as evidence of same-model self-leniency.

\paragraph{Contributions.}
The paper's contributions are both methodological and empirical:
\begin{itemize}[leftmargin=*,itemsep=1pt,topsep=2pt]
  \item \textbf{Protocol:} a controlled RAG meta-evaluation setup with hidden, localized source contradictions and blind judging against the original grounding.
  \item \textbf{Paired estimand:} an answer-level same-model contrast that compares matching and non-matching judges on the same candidate answer.
  \item \textbf{Empirical result:} no robust evidence that same-model judges broadly miss induced errors in their own model's answers.
  \item \textbf{Interpretation:} the remaining flagging gap reflects generator behavior and label/task mismatch, supported by targeted human evaluation.
\end{itemize}

\section{Related Work}

\paragraph{RAG and grounding evaluation.}
RAG combines parametric generation with retrieved non-parametric evidence \citep{lewis2020rag}. Automated RAG evaluators such as RAGAS and ARES score dimensions including faithfulness, context relevance, and answer correctness \citep{es2023ragas,saadfalcon2023ares}. Work on attribution, citation-supported generation, and factual consistency similarly treats grounding as claim/source alignment \citep{rashkin2023attribution,gao2023alce,honovich2022true,min2023factscore}. RAGTruth and FaithBench provide hallucination annotations or challenging factuality cases \citep{niu2023ragtruth,bao2024faithbench}. We use GaRAGe as the source dataset because it provides long-form answers and passage-level grounding over 2,366 questions and more than 35K passages \citep{sorodoc2025garage}. Our protocol is narrower than general RAG scoring: it creates paired original/perturbed grounding sets so the target contradiction is localized and hidden from the judge.

\paragraph{Contextual judges and refusal behavior.}
LLM judges are widely used to evaluate generated text and instruction-following outputs, including G-Eval, MT-Bench, Chatbot Arena, and Prometheus \citep{liu2023geval,zheng2023judging,kim2023prometheus}. Recent contextual judge benchmarks show that evaluating responses against context remains difficult. ContextualJudgeBench evaluates contextual judges with 2,000 response pairs and explicitly includes RAG-QA faithfulness and refusal behavior \citep{xu2025contextualjudgebench}. RAGferee trains RAG-centric contextual reward models for groundedness and appropriate refusals \citep{coman2025ragferee}. These studies motivate our behavior analysis: apparent false positives concentrate in refusals and conflict-aware outputs, not in ordinary committed answers.

\paragraph{Self-preference and controlled judge meta-evaluation.}
Prior work documents model-judge biases, including length, position, style, familiarity, and self-preference effects \citep{dubois2024length,shi2024position,liu2024narcissistic,wataoka2024selfpreference,ye2024justice}. A central difficulty is identification: a judge may favor its own model's output because of true quality, stylistic familiarity, or family-specific conventions rather than self-leniency. Recent work therefore argues for bias estimates that control for response quality or independent judgments \citep{spiliopoulou2025playfavorites,chen2025surface}. In fact-centric RAG, \citet{chen2025factrag} find no significant self-preference effect, suggesting that factual grounding may reduce the style-driven effects seen in open-ended preference tasks. Our work follows this identification-oriented line but moves it to a stricter source-grounded setting: we induce hidden contradictions in retrieved passages, generate natural answers from the perturbed grounding, and compare matching versus non-matching judges on the exact same candidate answer. This answer-paired design separates same-model judging effects from generator quality, refusal rates, and answer difficulty, while the behavior and human-evaluation analyses expose RAG-specific confounds that do not arise in ordinary preference scoring. REFLECT also uses controlled localized interventions for judge meta-evaluation, but studies evidence-based research-agent traces rather than RAG answer generation \citep{wang2026reflect}.

\section{Task and Data Construction}

\subsection{Base Dataset and Perturbation Design}

\begin{figure}[t]
  \centering
  \includegraphics[width=0.82\linewidth]{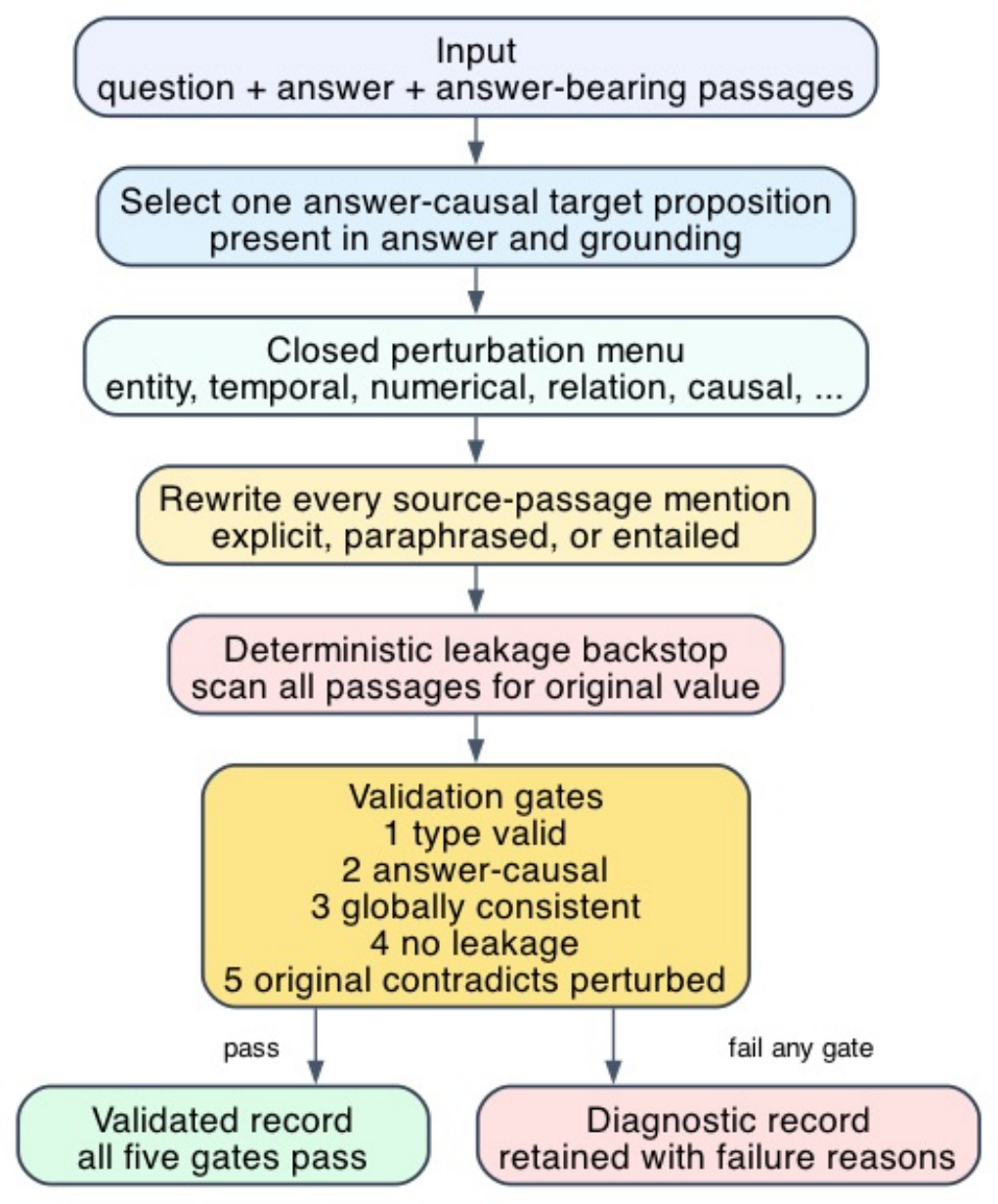}
  \caption{Perturbation and validation mechanics. A perturbation is valid for the primary analysis only if all five gates pass. Failed records are retained only as diagnostics.}
  \label{fig:perturb-validate}
\end{figure}

\begin{figure}[b]
  \centering
  \includegraphics[width=\linewidth]{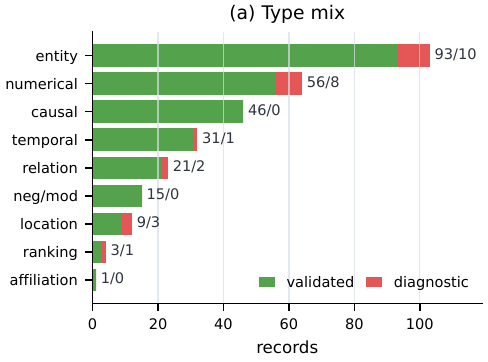}
  \caption{Perturbation-type composition of the 300 selected records, split by validated/diagnostic status.}
  \label{fig:data-composition}
\end{figure}

\begin{figure}[b]
  \centering
  \includegraphics[width=\linewidth]{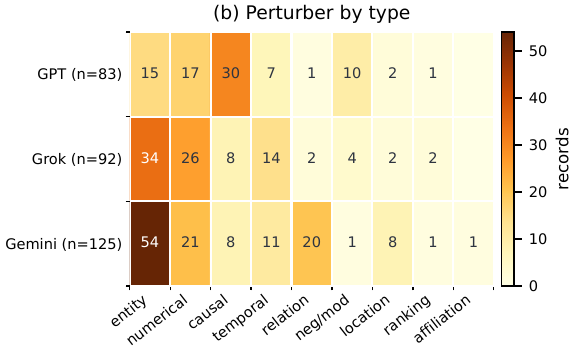}
  \caption{Perturber-by-type matrix for the 300 selected records; row totals give selected records per perturber.}
  \label{fig:data-composition-perturber}
\end{figure}

The pipeline starts from GaRAGe examples with a valid information-seeking question, a validated human-written answer, and at least two grounding passages labeled as answering the question. For each selected record, an LLM perturber receives the question, the human answer, and answer-bearing plus related grounding passages. The perturber must select exactly one answer-causal target proposition: a claim present in the answer and grounding such that changing it would change the answer to the question.

The closed perturbation menu includes entity substitutions, temporal shifts, numerical shifts, relation inversions, location and affiliation changes, ranking flips, causal changes, negation/modality changes, and multihop bridges. For every passage sent to the perturber, the output includes a full rewritten passage if the passage states or entails the selected claim, or a skip reason otherwise. The rewrite must remove the original value and any anchors that leak it. The selected 300-record pool is intentionally mixed rather than type-balanced; Figures~\ref{fig:data-composition} and~\ref{fig:data-composition-perturber} report the type mix and perturber mix; Appendices~\ref{app:dataset-selection}--\ref{app:perturbation} give details.

\paragraph{Perturbation examples.}
A numerical-shift record asks: ``How did Caitlin Clark's WNBA debut influence league attendance in 2024?'' The original answer-causal claim is that the debut led to a 48\% increase in overall league attendance. The perturber changes the value to 35\%, producing a passage that says the league saw a 35\% jump while preserving the rest of the passage. A faithful generator reading only the perturbed passage should answer with 35\%, but that answer is false relative to the original passage.

A relation-inversion record is more semantically disruptive. The original Zantac-trial claim says a jury returned a defense verdict, producing a win and rising stock prices. The perturbed claim says the jury returned a plaintiff verdict, producing a loss and plummeting stock prices. This illustrates why relation inversions require consistency edits around the target proposition rather than a single-token replacement.

\subsection{Validation and Diagnostic Records}

Each perturbation is validated by an LLM validation pass plus deterministic leakage checks; Figure~\ref{fig:perturb-validate} shows the five-gate flow. The five gates are type validity, answer causality, global context consistency, no original-answer leakage, and original contradicts perturbed. The final evaluation pool has 300 unique core records selected from perturbations proposed by GPT, Grok, and Gemini perturbers: 83, 92, and 125 selected records, respectively. Of these, 275 pass all five gates and 25 are failed-validation diagnostic records. We use the 275 validated records for every main result and report the 25 diagnostics only as a sensitivity analysis. Figure~\ref{fig:data-composition} shows that diagnostics are concentrated by count in entity and numerical records, with larger diagnostic shares in the smaller location and ranking strata.

\section{Generation, Labels, and Judging}

\subsection{RAG Generation from Perturbed Grounding}

Each generator receives the original question and the perturbed grounding passages. The prompt instructs the model to answer using only the provided passages, cite passage IDs supporting factual claims, and refuse if the evidence is insufficient or inconsistent. We run three generator models over all 300 records, yielding 900 intended outputs and 897 usable outputs; the prompt is in Appendix~\ref{app:generation}.

The box illustrates the central case: the generator follows the modified grounding and becomes false against the original passages.

\begin{center}
\setlength{\fboxsep}{2.5pt}
\fcolorbox{black!45}{black!3}{%
\begin{minipage}{\dimexpr\linewidth-2\fboxsep-2\fboxrule\relax}
\scriptsize
\newcommand{\genmark}[1]{\begingroup\setlength{\fboxsep}{0.5pt}\colorbox{yellow!25}{#1}\endgroup}
\newcommand{\genboxlabel}[2]{\begin{tabular}[t]{@{}l@{}}\textbf{#1}\\\textbf{#2}\end{tabular}}
\noindent\textbf{Context-Follow Generation Example}\par\vspace{3pt}
\setlength{\tabcolsep}{2.5pt}
\renewcommand{\arraystretch}{1.0}
\begin{tabularx}{\linewidth}{@{}L{1.35cm}Y@{}}
\textbf{Question} & How did Caitlin Clark's WNBA debut influence league attendance in 2024? \\\noalign{\vskip 5pt}
\genboxlabel{Value}{swap} & 48\% $\rightarrow$ 35\% \\\noalign{\vskip 5pt}
\genboxlabel{Original}{passage} & In her rookie WNBA season, the league saw a \genmark{48\% jump} in attendance from 2023. The league's total attendance of 2,353,735 was the highest in 22 seasons. \\\noalign{\vskip 5pt}
\genboxlabel{Modified}{passage} & In her rookie WNBA season, the league saw a \genmark{35\% jump} in attendance from 2023. The league's total attendance of 2,353,735 was the highest in 22 seasons. \\\noalign{\vskip 5pt}
\textbf{GPT answer} & Leaguewide attendance rose 35\% from 2023, reaching 2,353,735 total fans, the highest in 22 seasons. \\\noalign{\vskip 5pt}
\genboxlabel{Behavior}{label} & Context follow: the answer adopts the perturbed 35\% value, so it is false relative to the original grounding. \\
\end{tabularx}
\end{minipage}}
\end{center}

\subsection{Answer Labels and Blind Judging}

\begin{figure}[t]
  \centering
  \includegraphics[width=0.98\linewidth]{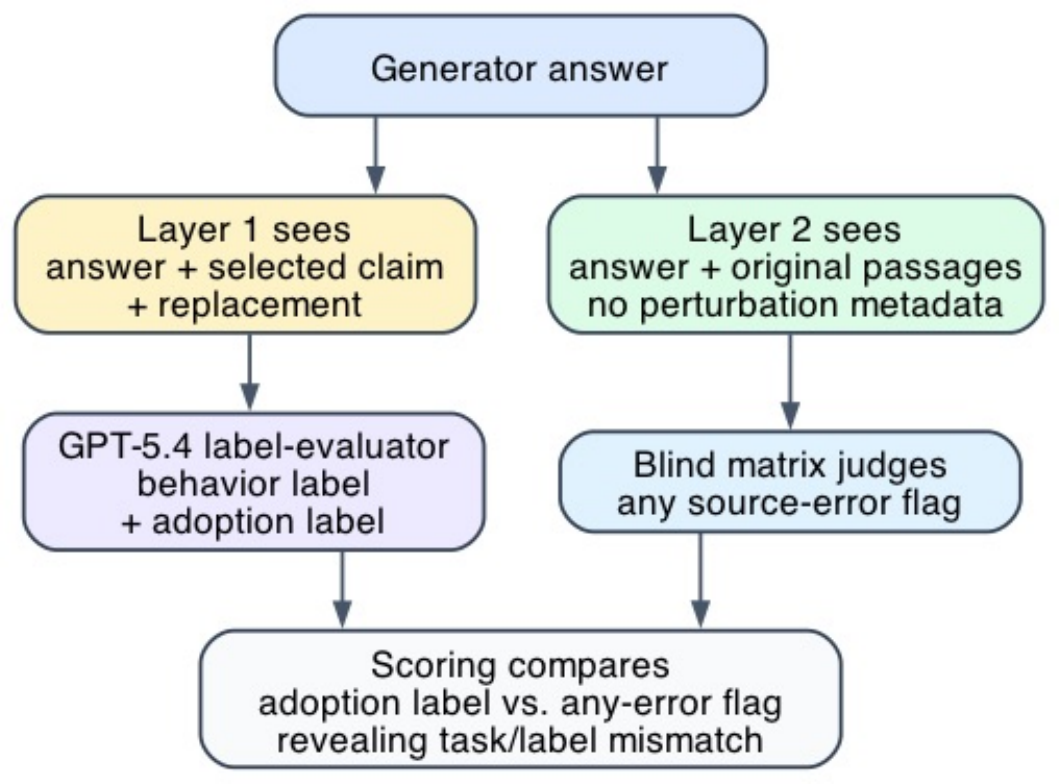}
  \caption{Two-layer setup. The induced-error adoption label comes from a GPT-5.4 label-evaluator with perturbation access; the matrix judges are blind and are asked for any source-relative factual error. This task/label mismatch is central to interpreting apparent false positives.}
  \label{fig:two-layer}
\end{figure}

Figure~\ref{fig:two-layer} shows the key separation: labels use perturbation metadata, while the matrix judges are blind to it.

After generation, we assign two answer-level labels before judging. The behavior label describes how an answer responds to the perturbed evidence. We use six generator-behavior labels:
\begin{itemize}[label={},leftmargin=0pt,itemsep=0pt,topsep=2pt,parsep=0pt]
  \item \textbf{Context follow:} asserts the perturbed value.
  \item \textbf{Memory override:} asserts the original value.
  \item \textbf{Both claims:} includes both original and perturbed values.
  \item \textbf{Conflict awareness:} reports inconsistent evidence.
  \item \textbf{Refusal/insuff.:} refuses or says the evidence is insufficient.
  \item \textbf{Unrelated/failed:} gives an off-task or unusable answer.
\end{itemize}
A separate LLM label-evaluation step assigns these labels using paraphrase-aware criteria; deterministic string matching is retained only as a diagnostic signal.

As Table~\ref{tab:gen-behaviors} shows, the share of non-context-following outputs differs by model: Grok refuses or declares insufficiency far more often than GPT, while GPT more often produces an answer that avoids the specific perturbed span. These behavior differences later explain much of the judge matrix.

\begin{table}[H]
\centering
\scriptsize
\begin{tabular}{lrrr}
\toprule
Behavior label & GPT & Grok & Gemini \\
\midrule
Context follow & 153 (51.2\%) & 141 (47.2\%) & 143 (47.8\%) \\
Memory override & 10 (3.3\%) & 8 (2.7\%) & 4 (1.3\%) \\
Both claims & 20 (6.7\%) & 15 (5.0\%) & 18 (6.0\%) \\
Conflict aware & 3 (1.0\%) & 12 (4.0\%) & 4 (1.3\%) \\
Refusal/insuff. & 16 (5.4\%) & 74 (24.7\%) & 54 (18.1\%) \\
Unrelated/failed & 97 (32.4\%) & 49 (16.4\%) & 76 (25.4\%) \\
\bottomrule
\end{tabular}
\caption{Generator behavior labels over 897/900 LLM-labeled outputs. Percentages use the labeled denominator for each generator (299).}
\label{tab:gen-behaviors}
\end{table}

\paragraph{Induced-error adoption labels and blind judging.}
The experiment has two LLM-evaluation layers. Layer 1 is a single GPT-5.4 label-evaluator that sees the answer and perturbation metadata and returns a behavior label plus \texttt{entails\_perturbed\_claim}. The induced-error adoption label is derived from that entailment flag: did the answer adopt the false value introduced by the perturbation? Layer 2 consists of the three crossed matrix judges. They see only the question, candidate answer, and original passages. They are blind to the perturbed grounding, original value, replacement value, generator behavior label, and induced-error adoption label.

For judging, each judge outputs a verdict, factual-error flag, wrong claim, source passage ID, explanation, and confidence; schemas are in Appendix~\ref{app:evaluation}. The main run uses one exact model endpoint per provider family: GPT-5.4, Grok 4.3, and Gemini 3.5 Flash. We use \emph{same-model} to mean the matched generator-judge cells (GPT--GPT, Grok--Grok, Gemini--Gemini), not a broader provider-family claim.

\section{Results}

\subsection{Validated Judge Matrix}

Table~\ref{tab:validated-matrix-main} shows the validated-only judge-by-generator matrix. Across cells, F1 ranges from 79.7\% to 87.9\%. The diagonal is not obviously inflated: mean diagonal F1 is 84.4 and mean off-diagonal F1 is 83.4. Grok-generated answers are the hardest column for every judge, showing that raw diagonal-vs-off-diagonal comparisons mix possible same-model effects with generator-column difficulty.

\begin{table*}[b]
\centering
\scriptsize
\begin{tabular*}{0.95\linewidth}{@{\extracolsep{\fill}}l ccc ccc ccc c@{}}
\toprule
 & \multicolumn{3}{c}{GPT answers} & \multicolumn{3}{c}{Grok answers} & \multicolumn{3}{c}{Gemini answers} & \\
\cmidrule(lr){2-4}\cmidrule(lr){5-7}\cmidrule(lr){8-10}
Judge & P & R & F1 & P & R & F1 & P & R & F1 & Row F1 \\
\midrule
GPT & \textbf{94.0} & \textbf{78.1} & \textbf{85.3} & 77.9 & 81.6 & 79.7 & 81.6 & 83.2 & 82.4 & 82.5 \\
Grok & 90.0 & 77.8 & 83.4 & \textbf{80.1} & \textbf{79.6} & \textbf{79.9} & 85.8 & 85.8 & 85.8 & 83.0 \\
Gemini & 89.2 & 86.4 & 87.8 & 78.1 & 85.0 & 81.4 & \textbf{85.4} & \textbf{90.6} & \textbf{87.9} & 85.7 \\
\midrule
Col F1 &  &  & 85.5 &  &  & 80.3 &  &  & 85.4 &  \\
\bottomrule
\end{tabular*}
\caption{Validated-only matrix (275 validated records; diagonal = same-model cells, bold). Per cell: precision/recall/F1 (\%). Mean diagonal F1 $=84.4$, mean off-diagonal F1 $=83.4$.}
\label{tab:validated-matrix-main}
\end{table*}

\subsection{Answer-Paired Same-Model Effects}

Every candidate answer is judged by all three judges, so we can compare judges on the same answer rather than comparing different generator columns. For answer $a$ generated by model $g$, the matching judge is $g$ and the two non-matching judges are the other models. Let $Y_{j,a}$ be judge $j$'s binary flag. The same-answer contrast is
\[
\Delta_a = Y_{g,a} - \frac{1}{2}\sum_{j \neq g}Y_{j,a}.
\]
A positive $\Delta_a$ means the matching judge flags more often than the other two judges on that answer; a negative value means it flags less often. We average this contrast three ways: over answers that adopted the induced false claim (recall $\Delta$), over answers that avoided that claim (avoided-claim flag $\Delta$), and over all answers (all-answer flag $\Delta$). Confidence intervals use a cluster bootstrap by core record ID with 10,000 replicates.

Table~\ref{tab:paired-effects} gives the paired estimates. The prespecified global recall effect is near zero ($-0.5$ pp, 95\% CI $[-2.7,+1.7]$), so the paired analysis does not support the hypothesis that same-model judges broadly miss induced errors in their own model's answers. The robust negative avoided-claim flag effect means that matching judges flag answers that avoided the induced false claim less often than the two non-matching judges. This should not be read as a lower false-alarm rate, because these answers can still contain other source-relative errors; Section~\ref{sec:human-evaluation} audits that mismatch. Appendices~\ref{app:old-vs-paired} and~\ref{app:sensitivity} give raw-vs-paired, behavior-sliced, and dataset-slice sensitivity checks.

\begin{table}[H]
\centering
\scriptsize
\renewcommand{\arraystretch}{1.08}
\begin{tabular*}{\linewidth}{@{\extracolsep{\fill}}lccc@{}}
\toprule
Set & \makecell{Recall\\$\Delta$} & \makecell{Avoided-claim\\flag $\Delta$} & \makecell{All-answer\\flag $\Delta$} \\
\midrule
\multicolumn{4}{@{}l}{\emph{Confirmatory}} \\
Global & -0.5 [-2.7,+1.7] & -4.3 [-6.6,-2.0] & -2.2 [-3.8,-0.6] \\
\midrule
\multicolumn{4}{@{}l}{\emph{Exploratory, by generator}} \\
GPT & -3.8 [-8.0,+0.3] & -6.7 [-11.4,-2.6] & -5.0 [-8.1,-2.0] \\
Grok & -3.7 [-8.7,+0.7] & -4.7 [-9.3,-0.4] & -4.2 [-7.5,-1.1] \\
Gemini & +6.0 [+2.9,+9.6] & -1.6 [-4.7,+1.6] & +2.6 [+0.2,+5.0] \\
\bottomrule
\end{tabular*}
\caption{Paired same-model effect: matching judge minus the mean of the two non-matching judges on the same answer. Each numeric cell is $\Delta$ [95\% cluster-bootstrap CI] in percentage points. Per-generator rows are exploratory and uncorrected.}
\label{tab:paired-effects}
\end{table}
\FloatBarrier

\subsection{Generator Behavior Explains the Matrix}

Table~\ref{tab:behavior-stratified} stratifies judge outcomes by generator behavior. Here TP/FP/FN/TN are computed against the induced-error adoption label: TP = adopted and flagged, FP = avoided but flagged, FN = adopted but unflagged, and TN = avoided and unflagged. Recall is concentrated in context-following answers: 1,025 of 1,139 true positives. Apparent false positives concentrate in conflict-aware and refusal/insufficient answers: 176 of 209. Memory override is a useful control: when the generator uses the original value, judges almost never flag an error.

\begin{table*}[t]
\centering
\scriptsize
\begin{tabular*}{0.95\linewidth}{@{\extracolsep{\fill}}l rr rrrr l@{}}
\toprule
Generator behavior & verdicts & answers & TP & FP & FN & TN & \makecell[l]{Headline rate\\{}[95\% CI]} \\
\midrule
context-follow & 1212 & 405 & 1025 & 0 & 187 & 0 & recall 84.6 [81.5,87.6] \\
both-claims & 135 & 45 & 88 & 0 & 44 & 3 & recall 66.7 [54.5,78.8] \\
memory-override & 42 & 14 & 0 & 0 & 0 & 42 & FPR 0.0 [0.0,7.1] \\
conflict-aware & 54 & 18 & 3 & 46 & 0 & 5 & FPR 90.2 [76.5,100.0] \\
refusal/insuff. & 400 & 134 & 23 & 130 & 1 & 246 & FPR 34.6 [27.4,42.2] \\
unrelated/failed & 606 & 202 & 0 & 25 & 0 & 581 & FPR 4.1 [2.1,6.3] \\
unlabeled & 9 & 3 & 0 & 8 & 0 & 1 & FPR 88.9 [66.7,100.0] \\
\bottomrule
\end{tabular*}
\caption{Judge outcomes pooled over all nine cells, stratified by LLM behavior labels. The reported rate is recall for answers that adopted the induced false claim and apparent false-positive rate otherwise.}
\label{tab:behavior-stratified}
\end{table*}

\subsection{Human Evaluation of Apparent False Positives}
\label{sec:human-evaluation}

The avoided-claim flagging gap in Table~\ref{tab:paired-effects} creates an interpretive question: when the induced-error adoption label calls a judge flag a false positive, is the judge actually wrong? In our setup, an apparent false positive is an answer that avoids the induced false claim but is still flagged by the judge. The label is narrower than the judge task: judges flag \emph{any} factual error against the original passages, so these answers can still contain another source-relative error.

We reviewed a targeted behavior-by-cell queue, capped at 12 cases per populated stratum, and adjudicated the first 88/153 cases in a shuffled, seed-fixed queue (mechanism audit, not prevalence estimate; codebook/counts: Appendices~\ref{app:audit-codebook},~\ref{app:sensitivity}). None of the 25 reviewed apparent false positives were genuine false alarms: 22 were valid alternate-error detections, 2 were adoption-label mistakes, and 1 was unclear.

\begin{center}
\setlength{\fboxsep}{2.5pt}
\fcolorbox{black!45}{black!3}{%
\begin{minipage}{\dimexpr\linewidth-2\fboxsep-2\fboxrule\relax}
\scriptsize
\newcommand{\swapmark}[1]{\begingroup\setlength{\fboxsep}{0.5pt}\colorbox{yellow!25}{#1}\endgroup}
\newcommand{\boxlabel}[2]{\begin{tabular}[t]{@{}l@{}}\textbf{#1}\\\textbf{#2}\end{tabular}}
\noindent\textbf{Apparent False Positive Example}\par\vspace{3pt}
\setlength{\tabcolsep}{2.5pt}
\renewcommand{\arraystretch}{1.0}
\begin{tabularx}{\linewidth}{@{}L{1.35cm}Y@{}}
\textbf{Question} & How have recent SEC changes affected operating lease reporting practices? \\\noalign{\vskip 5pt}
\boxlabel{Induced}{claim} & proactive initiatives targeting compliance failures $\rightarrow$ detailed audits of financial reporting accuracy \\\noalign{\vskip 5pt}
\boxlabel{Original}{passage} & SEC's fiscal year 2024 report saw unprecedented financial penalties, \swapmark{proactive initiatives} \swapmark{targeting compliance failures}, and heightened scrutiny of executives and gatekeepers. \\\noalign{\vskip 5pt}
\boxlabel{Modified}{passage} & SEC's fiscal year 2024 report saw unprecedented financial penalties, \swapmark{detailed audits} \swapmark{of financial} \swapmark{reporting accuracy}, and heightened scrutiny of executives and gatekeepers. \\\noalign{\vskip 5pt}
\textbf{GPT answer} & Avoids the induced SEC claim, but says recent changes made operating leases more balance-sheet focused and cites ASC 842. \\\noalign{\vskip 5pt}
\textbf{Grok flag} & Valid: the source attributes ASC 842 to FASB, not the SEC, and no source links SEC changes to those lease rules. \\\noalign{\vskip 5pt}
\textbf{Why FP?} & The adoption label says ``avoided induced claim''; the judge task says ``contains another source-relative error.'' Human evaluation treats it as a valid alternate-error detection. \\
\end{tabularx}
\end{minipage}}
\end{center}

This supports a different interpretation from self-leniency. The robust avoided-claim gap is better read as a label/task mismatch: the adoption label asks whether the answer took up the induced false claim, while the judge is allowed to catch any source-relative error. The boxed SEC/FASB case shows the distinction. The answer avoids the induced SEC claim, but it makes a separate source-grounding error by attributing an ASC 842 lease-accounting effect to recent SEC changes. Thus, under the any-error judge task, this is a valid detection; under the narrower induced-error adoption label, it is counted as an apparent FP. Table~\ref{tab:audit-summary} summarizes the reviewed cases.

\noindent\begin{minipage}{\linewidth}
\centering
\scriptsize
\begin{tabularx}{\linewidth}{@{}L{1.35cm}rL{1.75cm}Y@{}}
\toprule
\makecell[l]{Automatic\\case} & Reviewed & Human finding & Interpretation \\
\midrule
\makecell[l]{Apparent\\FP} & 25 & 0 genuine false alarms & mostly alternate source errors or adoption-label mistakes \\
TN & 31 & 31 genuine TN & no missed errors found in this control set \\
TP & 23 & 21 genuine TP & recall labels mostly valid \\
FN & 9 & 6 genuine misses & some adoption labels too broad \\
\bottomrule
\end{tabularx}
\captionsetup{hypcap=false,skip=2pt}
\captionof{table}{Human evaluation summary. Rule-of-three upper bounds are about 12\% for genuine false alarms among reviewed apparent FPs and about 10\% for missed errors among reviewed TNs.}
\label{tab:audit-summary}
\end{minipage}
\vspace{-1.2\baselineskip}

\subsection{Localization Remains Strong}

\begin{figure}[H]
  \centering
  \includegraphics[width=\linewidth]{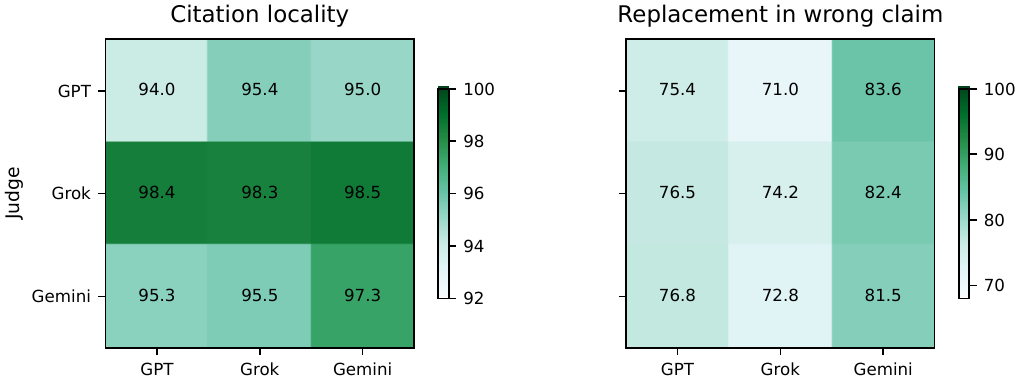}
  \caption{Localization metrics from the raw judge verdicts. Passage-level localization is consistently high; exact replacement containment in the wrong-claim text is lower because judges often paraphrase the erroneous claim.}
  \label{fig:localization}
\end{figure}

For downstream debugging, a judge is more useful when it points to the contradicted source. Figure~\ref{fig:localization} summarizes full-set localization metrics; Appendix~\ref{app:localization-table} gives cell-level values. When judges flag an error, their supporting passage lies in the known modified passage set 94.0--98.5\% of the time. Wrong-claim strings contain the replacement value 71.0--83.6\% of the time, reflecting paraphrase rather than passage-localization failure.

\FloatBarrier

\section{Discussion}

\paragraph{Same-model effects are identification claims.}
The full judge-by-generator matrix is valuable because it reveals heterogeneous generator difficulty and judge behavior. But a raw diagonal-vs-off-diagonal contrast compares different answers, so it cannot by itself distinguish self-leniency from answer quality, style, refusal rates, or cell difficulty. This mirrors the identification problem emphasized in prior self-preference work. The paired estimand changes the claim being tested: after holding the candidate answer fixed and comparing matching versus non-matching judges on that same answer, the global recall effect is near zero.

\paragraph{Fact-centric RAG changes the failure mode.}
The near-zero paired recall result is consistent with recent evidence that factual grounding can reduce broad style-driven self-preference in RAG. In this setting, the judge is not choosing which response it likes better; it is checking whether the response contradicts the source passages. That does not make the measurement problem disappear. Instead, the main ambiguity moves from generic preference bias to RAG-specific behavior and label/task alignment.

\paragraph{Behavior and labels determine the apparent flagging gap.}
The robust avoided-claim flagging gap is not evidence that judges simply over- or under-flag same-model outputs. The automatic adoption label records whether the answer used the induced false claim, while the judge task asks for any factual error against the original grounding. Context-following answers therefore create ordinary induced-error recall cases. Refusals, hedges, and conflict-aware answers can avoid the induced claim while still making another source-relative error. The human evaluation found no genuine false alarms among 25 reviewed apparent FPs, which supports interpreting the gap as a label/task mismatch rather than as evidence of same-model self-leniency.

\paragraph{Implications for practice.}
LLM judges can be useful source-grounding evaluators: the matrix shows strong precision/recall and high passage localization. But evaluations that reuse a model family as both generator and judge should report the full matrix, answer-paired same-vs-cross contrasts, and behavior strata. Apparent false positives should also be adjudicated into induced-error detection, alternate valid error detection, label error, and genuine false alarm. Aggregated same-model metrics are not enough, especially when refusal behavior differs by generator.

\section{Conclusion}

We introduced \method, a controlled protocol for testing same-model LLM judging in source-grounded RAG. The design induces one hidden answer-causal contradiction in the grounding, elicits answers from perturbed evidence, and judges those answers against the original evidence in a crossed judge-by-generator matrix. This makes the central estimand answer-paired: for the same candidate answer, does the matching judge behave differently from non-matching judges?

On 275 validated records, the answer-paired recall effect is near zero, despite heterogeneous generator behavior and raw cell differences. The remaining robust gap, lower matching-judge flagging on answers that avoided the induced claim, does not support a simple false-alarm interpretation under the any-error judge task. A targeted human evaluation finds that reviewed apparent false positives are alternate valid source-error detections, adoption-label mistakes, or unclear cases rather than genuine false alarms.

The broader lesson is that same-model RAG judging is an identification problem, not a yes/no prohibition. Full judge-by-generator matrices are useful descriptively, but claims about self-leniency require answer-level pairing, behavior stratification, and explicit alignment between the ground-truth label and the error task assigned to the judge.

\section*{Reproducibility and Artifacts}

The artifact repository contains the frozen 300-record manifest and selected record IDs, perturbation metadata, original/perturbed claim pairs, validation-gate outputs, prompts, JSON schemas, generator outputs, behavior/adoption labels, judge verdicts, paired-analysis inputs, human-review codebook, seed-fixed review queue, and analysis scripts: \url{https://anonymous.4open.science/r/eval-pair-matrix-emnlp-2026/}. Where source-passage redistribution is restricted by upstream licensing, the release provides passage IDs and derived perturbation metadata sufficient to reproduce the reported analyses from GaRAGe.

\section*{Limitations}

The dataset is controlled and moderate in scale: 275 validated records and 25 failed-validation diagnostics. We evaluate one exact model endpoint per provider, so the design cannot separate model identity from provider-family or style-familiarity effects. The behavior labels and induced-error adoption labels come from a single GPT-5.4 label-evaluator rather than independent human annotation; the human evaluation found low but nonzero internal contradiction and about 8\% adoption-label issues in reviewed refusal/conflict cases. The human evaluation is targeted and single-adjudicator, not a double-annotated study with agreement statistics, and is used to diagnose the apparent-FP mechanism rather than estimate population false-alarm prevalence. Judge calls are single-shot, so stochastic flip rates are not estimated. Finally, the strict source-grounding stance used in the human evaluation is defensible but consequential: an alternative adoption-only interpretation would classify some conflict/refusal flags as artifacts rather than alternate errors, though not as genuine false alarms.

\section*{Ethics Statement}

The work creates controlled factual perturbations for evaluation. Perturbed passages should not be presented as factual information, and any release should preserve provenance, validation metadata, and original/perturbed labels. The benchmark does not evaluate people or make consequential decisions about individuals. The main risk is misuse of the perturbation pipeline to produce misleading text; the mitigation is to release it only as a traceable evaluation artifact with clear documentation that the corrupted passages are synthetic counterfactuals.

%% file: appendix_content.tex
\clearpage
\appendix

\section*{Appendix}

This appendix follows the same order as the main paper so that a reader can move from the construction of the benchmark to the final sensitivity checks without changing mental frames. Section~\ref{app:dataset-selection} documents how the 300-record pool was selected. Section~\ref{app:perturbation} shows how records were rewritten and validated. Section~\ref{app:generation} records the generator prompt, behavior examples, and generator-side diagnostics. Section~\ref{app:evaluation} gives the judge and label-evaluation details, including the human-evaluation codebook. Section~\ref{app:results} then collects supporting result tables, sensitivity analyses, and reproducibility details.

\section{Dataset Selection}
\label{app:dataset-selection}

The experiment starts from GaRAGe and freezes a controlled 300-record pool before perturbation, generation, and judging. Table~\ref{tab:dataset-funnel} summarizes the selection path. The initial filter keeps examples with a valid information-seeking question, no false premise, a validated human answer, and at least two grounding passages labeled as answering the question. The selected pool is then stratified across question popularity, complexity, and category.

\begin{table}[H]
\centering
\scriptsize
\begin{tabularx}{\linewidth}{@{}L{1.45cm}Y r@{}}
\toprule
Stage & Rule or source & Records \\
\midrule
Raw GaRAGe train split & Full local source file used for sampling & 2,366 \\
Eligible after filter & Valid, seeking question; no false premise; validated answer; at least two answer-bearing passages & 1,723 \\
Base selection & Round-robin stratified sample over popularity, complexity, and category; seed 17 & 300 \\
Final multi-provider pool & Selected from GPT, Grok, and Gemini perturbation/validation runs & 300 \\
Primary analysis set & All five validation gates pass & 275 \\
Diagnostic set & One or more validation gates fail; retained only for diagnostics and sensitivity & 25 \\
\bottomrule
\end{tabularx}
\caption{Dataset-selection funnel. The final 300 records are frozen before generator and judge runs; the main paper reports the 275 fully validated records as the primary set.}
\label{tab:dataset-funnel}
\end{table}

\begin{table}[H]
\centering
\scriptsize
\begin{tabularx}{\linewidth}{@{}Y r Y r@{}}
\toprule
Popularity & $n$ & Complexity & $n$ \\
\midrule
Tail & 129 & Multi-hop & 83 \\
Torso & 99 & Post-processing heavy & 65 \\
Head & 72 & Set & 49 \\
 & & Simple w. condition & 42 \\
 & & Comparison & 40 \\
 & & Simple & 13 \\
 & & Aggregation & 8 \\
\bottomrule
\end{tabularx}
\caption{Question mix in the frozen 300-record pool. The most frequent top-level categories are Law \& Government, Arts \& Entertainment, Sports, Finance, and Business \& Industrial.}
\label{tab:dataset-mix}
\end{table}

The example below illustrates the kind of record retained by this filter: the question is grounded by multiple passages, has a validated answer, and is neither trivia-like nor underspecified.

\begin{center}
\setlength{\fboxsep}{2.5pt}
\fcolorbox{black!45}{black!3}{%
\begin{minipage}{\dimexpr\linewidth-2\fboxsep-2\fboxrule\relax}
\scriptsize
\setlength{\tabcolsep}{2pt}
\renewcommand{\arraystretch}{1.05}
\begin{tabularx}{\linewidth}{@{}L{1.35cm}Y@{}}
\textbf{Example} & How did Caitlin Clark's WNBA debut influence league attendance in 2024? \\\noalign{\vskip 4pt}
\textbf{Metadata} & Sports; torso-popularity; multi-hop; slow-changing. \\\noalign{\vskip 4pt}
\textbf{Grounding} & Eleven answer-bearing passages were available; the later perturbation modified passages 1, 3, and 12. \\\noalign{\vskip 4pt}
\textbf{Why kept} & The question is answer-seeking, the human answer is validated, and multiple passages directly support the answer. \\
\end{tabularx}
\end{minipage}}
\end{center}

\FloatBarrier

\section{Perturbation}
\label{app:perturbation}

Each selected record is sent to an LLM perturber with the question, human answer, answer-bearing passages, and related grounding passages. The perturber must choose one answer-causal value and rewrite every passage that states or entails that value. The prompt excerpt below is taken from the implementation.

\begin{lstlisting}[basicstyle=\ttfamily\tiny]
Input: question, human answer, grounding passages.
Task: pick one atomic value present in both answer and grounding.
The value must be answer-causal: a faithful answer to the rewritten
grounding should materially differ from the original answer.
Choose one perturbation type from a closed menu.
Rewrite every passage mention consistently; remove the original value
and contextual anchors that would reveal it.
Return JSON only: perturbation_type, original_value, perturbed_value,
atomic_claim_original, atomic_claim_perturbed, plausibility,
deducibility_note, and one doc_modifications entry per passage.
\end{lstlisting}

The validation step has two roles. It removes records where the rewrite is not a clean counterfactual, and it keeps diagnostic failures visible so readers can see what kinds of examples were excluded from the primary analysis.

\begin{table}[H]
\centering
\scriptsize
\begin{tabularx}{\linewidth}{@{}L{1.75cm}Y@{}}
\toprule
Validation gate & What it checks \\
\midrule
Type validity & The rewrite preserves grammar and entity/value type. \\
Answer causality & A faithful answer would change under the perturbed claim. \\
Global consistency & Every rewritten passage supports the perturbed claim and none still support the original. \\
No leakage & The original value and unique contextual anchors disappear from perturbed passages. \\
Contradiction & Original passages support the original claim and contradict the perturbed claim. \\
\bottomrule
\end{tabularx}
\caption{Five validation gates used to decide whether a perturbation enters the primary analysis set.}
\label{tab:validation-gates}
\end{table}

\begin{table}[H]
\centering
\scriptsize
\begin{tabularx}{\linewidth}{@{}Y rrr@{}}
\toprule
Perturbation type & Total & Valid. & Diag. \\
\midrule
Entity substitution & 103 & 93 & 10 \\
Numerical shift & 64 & 56 & 8 \\
Causal change & 46 & 46 & 0 \\
Temporal shift & 32 & 31 & 1 \\
Relation inversion & 23 & 21 & 2 \\
Negation/modality & 15 & 15 & 0 \\
Location change & 12 & 9 & 3 \\
Ranking flip & 4 & 3 & 1 \\
Affiliation change & 1 & 1 & 0 \\
\bottomrule
\end{tabularx}
\caption{Perturbation-type distribution in the 300-record pool. Counts are computed before generator runs; the primary analysis uses the validated subset.}
\label{tab:perturb-type-counts}
\end{table}

\begin{table}[H]
\centering
\scriptsize
\begin{tabularx}{\linewidth}{@{}Y r Y r@{}}
\toprule
Modified passages & Records & Modified passages & Records \\
\midrule
1 & 108 & 5 & 21 \\
2 & 63 & 6--8 & 34 \\
3 & 39 & 9+ & 15 \\
4 & 20 & & \\
\bottomrule
\end{tabularx}
\caption{How many grounding passages were rewritten per selected record. The median is 2, the mean is 3.08, and the maximum is 14.}
\label{tab:modified-passage-counts}
\end{table}

The next table makes the validation rules concrete. The first two rows are representative validated rewrites; the third is a diagnostic failure retained only to show why failed-validation records are separated from the main conclusions.

\begin{table}[H]
\centering
\scriptsize
\setlength{\tabcolsep}{2.5pt}
\renewcommand{\arraystretch}{1.12}
\begin{tabularx}{\linewidth}{@{}L{1.25cm}Y@{}}
\toprule
Example & Perturbation details \\
\midrule
numeric & \textbf{Type/value:} numerical shift; 48\% $\rightarrow$ 35\%. \newline
\textbf{Question:} How did Caitlin Clark's WNBA debut influence league attendance in 2024? \newline
\textbf{Rewrite:} In her rookie WNBA season, the league saw a 35\% jump in attendance from 2023. \newline
\textbf{Gate outcome:} all five gates pass. \\
relation & \textbf{Type/value:} relation inversion; defense verdict $\rightarrow$ plaintiff verdict. \newline
\textbf{Question:} What were the outcomes of the initial Zantac trials involving Abbott and Mead Johnson? \newline
\textbf{Rewrite:} The plaintiff verdict in Whitfield v. Abbott is a devastating loss for Abbott and Reckitt. \newline
\textbf{Gate outcome:} all five gates pass. \\
diagnostic & \textbf{Type/value:} numerical shift; two $\rightarrow$ seventeen. \newline
\textbf{Question:} What are the expected benefits and risks of Ethereum 2.0 for long-term investors? \newline
\textbf{Rewrite:} Ethereum's upcoming Pectra upgrade in 2025 is met with optimism, as seventeen out of its 19 upgrades historically generated notable price impacts. \newline
\textbf{Gate outcome:} fails type validity, answer causality, and leakage. \\
\bottomrule
\end{tabularx}
\caption{Representative perturbation and validation cases. The diagnostic example shows why the 25 failed-validation records are not used for the main conclusions.}
\label{tab:perturb-examples}
\end{table}

\FloatBarrier

\section{Generation}
\label{app:generation}

Generators receive the original question and the perturbed grounding only. They are asked to answer from the passages, cite passage IDs, and refuse when the evidence is insufficient or internally inconsistent.

\begin{lstlisting}[basicstyle=\ttfamily\tiny]
You answer a question using ONLY the provided grounding passages.
Base every factual claim on the passages.
Cite passage IDs using markers such as [3] or [3][7].
If the passages are insufficient, internally inconsistent, or you
cannot find the answer, set is_refusal=true and explain briefly.
Return JSON: answer, cited_passage_ids, is_refusal, notes.
\end{lstlisting}

After generation, a separate label-evaluator assigns the behavior label. Table~\ref{tab:generation-examples} shows four cases where a strict string match would be too brittle: answers can paraphrase the replacement, revert to background knowledge, refuse, or explicitly report inconsistent evidence.

\begin{table}[H]
\centering
\scriptsize
\setlength{\tabcolsep}{2.5pt}
\renewcommand{\arraystretch}{1.12}
\begin{tabularx}{\linewidth}{@{}L{1.25cm}Y@{}}
\toprule
Behavior & Generation case \\
\midrule
context follow & \textbf{Question:} How did Caitlin Clark's WNBA debut influence league attendance in 2024? \newline
\textbf{Value change:} 48\% $\rightarrow$ 35\%. \newline
\textbf{Answer snippet:} Caitlin Clark's 2024 WNBA debut/arrival was associated with a major jump in league attendance; leaguewide attendance rose 35\% from 2023, reaching 2,353,735 total fans. \newline
\textbf{Rationale:} states 35\%, matching the perturbed passages and not the original 48\%. \\
memory override & \textbf{Question:} How does Loschmidt's paradox challenge Boltzmann's H theorem? \newline
\textbf{Value change:} reversible $\rightarrow$ irreversible. \newline
\textbf{Answer snippet:} Loschmidt's paradox challenges Boltzmann's H-theorem by arguing that one should not derive irreversible entropy increase from time-symmetric or microscopically reversible dynamics. \newline
\textbf{Rationale:} states the standard formulation, contradicting the altered documents. \\
refusal & \textbf{Question:} How has Letsile Tebogo's cultural heritage influenced his athletic career? \newline
\textbf{Value change:} Botswana $\rightarrow$ Ghana. \newline
\textbf{Answer snippet:} The provided passages do not contain sufficient information on how Letsile Tebogo's cultural heritage has influenced his athletic career. \newline
\textbf{Rationale:} declines to assert an influence because the passages are insufficient. \\
conflict aware & \textbf{Question:} What impact has National Dog Month had on dog adoption rates? \newline
\textbf{Value change:} promotes $\rightarrow$ discourages adoption. \newline
\textbf{Answer snippet:} The passages do not provide a clear adoption-rate impact; they say National Dog Month promotes adoption while other passages inconsistently claim it discourages adoption. \newline
\textbf{Rationale:} explicitly reports inconsistent evidence. \\
\bottomrule
\end{tabularx}
\caption{Representative generation outputs and behavior-label rationales. These examples show why paraphrase-aware LLM labels were used instead of strict string matching.}
\label{tab:generation-examples}
\end{table}

\FloatBarrier

\begin{table}[H]
\centering
\scriptsize
\begin{tabularx}{\linewidth}{@{}Y ccc@{}}
\toprule
Generator & \makecell{LLM-labeled\\refusals} & \makecell{Self-flagged\\refusals} & \makecell{Context-follow\\cites modified} \\
\midrule
GPT & 16 & 19 & 152/153 \\
Grok & 74 & 90 & 141/141 \\
Gemini & 54 & 57 & 143/143 \\
\bottomrule
\end{tabularx}
\caption{Additional generation diagnostics. Refusal rates differ sharply by generator, while context-following answers almost always cite a modified passage.}
\label{tab:generation-diagnostics}
\end{table}

\begin{figure}[H]
  \centering
  \includegraphics[width=0.95\linewidth]{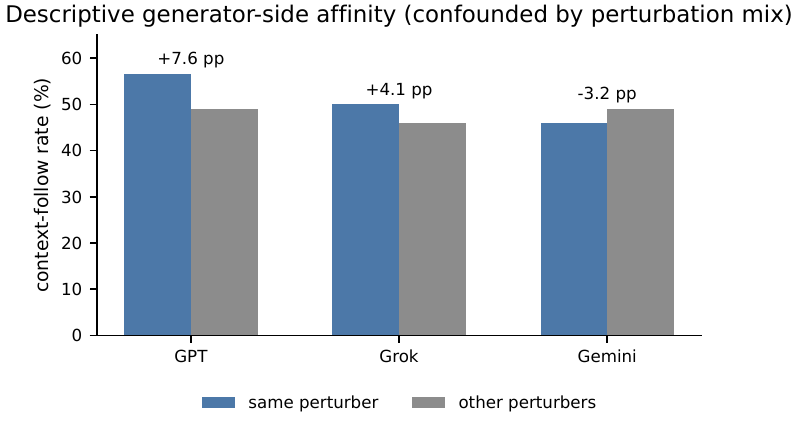}
  \caption{Generator-side self-affinity, retained as descriptive only. Same-perturber context-follow rates are confounded by provider-specific perturbation type preferences.}
  \label{fig:self-affinity}
\end{figure}

\begin{figure}[H]
  \centering
  \includegraphics[width=0.80\linewidth]{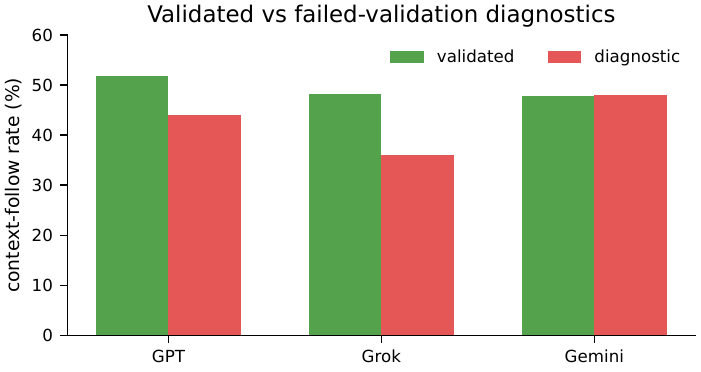}
  \caption{Context-following on validated versus diagnostic records. Diagnostic records are excluded from the primary analysis but still show meaningful generation behavior.}
  \label{fig:validated-vs-gap}
\end{figure}

\begin{figure}[H]
  \centering
  \includegraphics[width=0.80\linewidth]{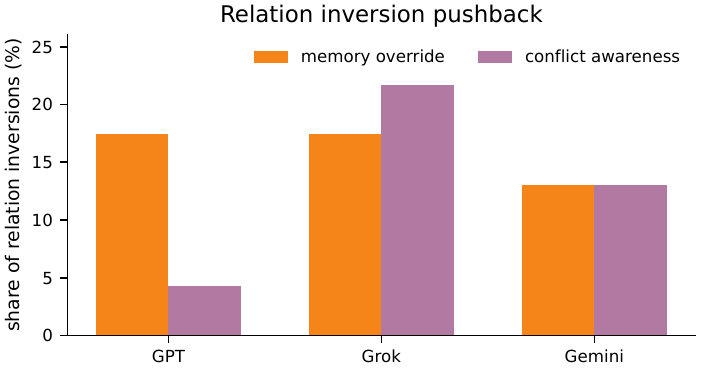}
  \caption{Relation-inversion pushback. Relation inversions elicit more memory override and conflict awareness than simpler substitutions.}
  \label{fig:relation-pushback}
\end{figure}

\section{Evaluation}
\label{app:evaluation}

The experiment has two evaluation layers. The label-evaluator sees perturbation metadata and assigns behavior and induced-error adoption labels. The matrix judges are blind: they see only the question, candidate answer, and original grounding passages.

\begin{lstlisting}[basicstyle=\ttfamily\tiny]
Label-evaluator input: question, original value, perturbed value,
original claim, perturbed claim, modified passages, generator answer.
Output: behavior_label, entails_original_claim,
entails_perturbed_claim, rationale, confidence.

Judge input: question, answer, original source passages.
Output: verdict, contains_factual_error, wrong_claim,
supporting_source_passage_id, explanation, confidence.
\end{lstlisting}

\begin{table}[H]
\centering
\scriptsize
\begin{tabularx}{\linewidth}{@{}L{1.4cm}ccY@{}}
\toprule
Cell & \makecell{Adopted\\induced claim?} & \makecell{Judge\\flagged?} & Interpretation \\
\midrule
TP & yes & yes & Judge caught an answer that adopted the induced false claim. \\
FP & no & yes & Apparent false positive: the answer avoided the induced claim but may contain another source error. \\
FN & yes & no & Judge missed an answer that adopted the induced false claim. \\
TN & no & no & Answer avoided the induced claim and judge did not flag an error. \\
\bottomrule
\end{tabularx}
\caption{Confusion-cell definitions used in the matrix. These cells are computed against the induced-error adoption label, while judges are asked to flag any source-relative factual error.}
\label{tab:cell-definitions}
\end{table}

\subsection{Human Evaluation Codebook}
\label{app:audit-codebook}

The targeted human evaluation records two judgments per case. The first checks whether the automatic induced-error adoption label is correct. The second asks whether the judge's flag/no-flag decision is correct against the original passages under a strict source-grounding stance. We adjudicated the first 88 completed cases in a shuffled, seed-fixed review queue before submission; all sampling strata and unreviewed cases are retained in the released queue. Because the audit is targeted rather than prevalence-estimating, it diagnoses the apparent-FP mechanism rather than estimating a population false-alarm rate.

\begin{table}[H]
\centering
\scriptsize
\begin{tabularx}{\linewidth}{@{}L{1.4cm}Y@{}}
\toprule
Call pattern & Derived outcome and interpretation \\
\midrule
TP, label ok, judge ok & \textbf{Genuine TP:} judge caught a valid source error in an answer that adopted the induced false claim. \\
TP, label ok, judge not ok & \textbf{Right verdict, wrong reason:} flag is correct but rationale/localization is not. \\
FP, label ok, judge ok & \textbf{Alternate error:} answer avoids the induced false claim but contains another valid source error. \\
FP, label ok, judge not ok & \textbf{Genuine false alarm:} judge flagged without a source error. \\
FP, label not ok & \textbf{Adoption-label mistake:} answer actually adopted the induced false claim. \\
FN, label ok, judge not ok & \textbf{Genuine miss:} induced error was missed. \\
FN, label not ok & \textbf{Adoption label too broad:} answer did not actually adopt the induced false claim. \\
TN, label ok, judge ok & \textbf{Genuine TN:} no induced error and no other source error found. \\
TN, judge not ok & \textbf{Missed error:} judge stayed silent despite a source error. \\
\bottomrule
\end{tabularx}
\caption{Human evaluation derivation rules. Apparent FPs in the automatic matrix can be false alarms, alternate valid source-error detections, or induced-error adoption-label mistakes.}
\label{tab:audit-codebook}
\end{table}

\begin{table}[H]
\centering
\scriptsize
\begin{tabularx}{\linewidth}{@{}Y cccc@{}}
\toprule
Behavior & TP & FP & FN & TN \\
\midrule
Context follow & 8/12 & -- & 9/12 & -- \\
Both claims & 0/12 & -- & 0/12 & 0/3 \\
Memory override & -- & -- & -- & 10/12 \\
Conflict awareness & 3/3 & 7/12 & -- & 0/5 \\
Refusal/insuff. & 12/12 & 8/12 & 0/1 & 10/12 \\
Unrelated/failed & -- & 10/12 & -- & 11/12 \\
Unlabeled & -- & 0/8 & -- & 0/1 \\
\bottomrule
\end{tabularx}
\caption{Human-evaluation review coverage by behavior and confusion cell, reported as completed adjudications over sampled queue entries. The queue sampled at most 12 cases per populated stratum.}
\label{tab:human-review-coverage}
\end{table}

The examples in Table~\ref{tab:judge-examples} illustrate why the matrix needs both automatic labels and human review. In particular, an apparent FP can be a real judge detection if the answer avoided the induced claim but introduced a different source-relative error.

\begin{table}[H]
\centering
\scriptsize
\setlength{\tabcolsep}{2.5pt}
\renewcommand{\arraystretch}{0.96}
\begin{tabularx}{\linewidth}{@{}L{1.15cm}Y@{}}
\toprule
Example & Judge case \\
\midrule
TP same & \textbf{Cell:} Gemini judge on Gemini answer; adoption label=true; judge flagged=true. \newline
\textbf{Question:} How did Caitlin Clark's WNBA debut influence league attendance in 2024? \newline
\textbf{Wrong claim:} overall league attendance jumped by 35\% compared to 2023. \newline
\textbf{Judge explanation:} the answer claims 35\%, while the original source states 48\%. \\
FN & \textbf{Cell:} GPT judge on GPT answer; adoption label=true; judge flagged=false. \newline
\textbf{Question:} How does RL contribute to personalized learning experiences in education? \newline
\textbf{Wrong claim:} --. \newline
\textbf{Judge explanation:} the judge says the answer is supported by the passages, missing the induced replacement. \\
Apparent FP & \textbf{Cell:} Grok judge on GPT answer; adoption label=false; judge flagged=true. \newline
\textbf{Question:} How have recent SEC changes affected operating lease reporting practices? \newline
\textbf{Wrong claim:} under ASC 842, companies must recognize most leases on the balance sheet. \newline
\textbf{Judge explanation:} Passage 2 attributes ASC 842 to FASB rather than SEC, and no source links SEC changes to the specific balance-sheet recognition requirement. \\
TN & \textbf{Cell:} GPT judge on Grok answer; adoption label=false; judge flagged=false. \newline
\textbf{Question:} How has Letsile Tebogo's cultural heritage influenced his athletic career? \newline
\textbf{Wrong claim:} --. \newline
\textbf{Judge explanation:} the passages mention Tebogo and Botswana's sports culture but do not explain how his cultural heritage influenced his career. \\
\bottomrule
\end{tabularx}
\caption{Representative judge outcomes. The apparent FP is a key example: the answer avoids the induced false claim but appears to be a valid alternate source-error detection.}
\label{tab:judge-examples}
\end{table}

\FloatBarrier
\newpage

\section{Results}
\label{app:results}

The result appendix separates exploratory summaries from the paired estimands used in the main conclusions. We first show how the older raw diagonal contrast differs from the answer-paired contrast, then provide full-set counts and sensitivity checks.

\subsection{Raw Diagonal Signatures Versus Paired Contrasts}
\label{app:old-vs-paired}

\begin{figure}[H]
  \centering
  \includegraphics[width=0.98\linewidth]{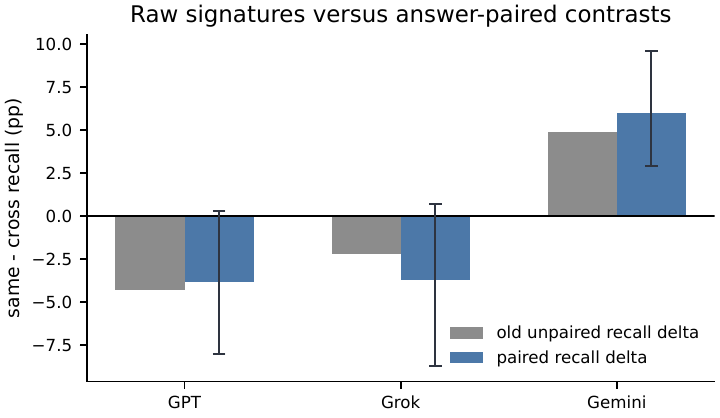}
  \caption{Raw diagonal recall deltas versus answer-paired recall deltas on the validated set. The paired analysis preserves some exploratory provider-level patterns but removes the basis for a broad self-leniency claim.}
  \label{fig:old-vs-paired}
\end{figure}

The raw diagonal contrast compares a judge's diagonal cell against that judge's two cross-generator cells. It is retained only as a descriptive diagnostic because it compares different answers. The paired contrast compares the matching judge with non-matching judges on the same answer.

\FloatBarrier

\subsection{Full-Set Judge Metrics}
\label{app:full-metrics}

The full 300-record matrix below is included for completeness. It differs from the main paper's validated-only matrix because it includes the 25 diagnostic records; the agreement between the full and validated slices is summarized in Section~\ref{app:sensitivity}.

\begin{table*}[t]
\centering
\small
\begin{tabular*}{\textwidth}{@{\extracolsep{\fill}}llrrrrrrrrr@{}}
\toprule
Judge & Gen. & $n$ & Adopted & TP & FP & FN & TN & Prec. & Rec. & F1 \\
\midrule
GPT & GPT & 297 & 173 & 134 & 9 & 39 & 115 & 93.7 & 77.5 & 84.8 \\
GPT & Grok & 299 & 159 & 131 & 38 & 28 & 102 & 77.5 & 82.4 & 79.9 \\
GPT & Gemini & 296 & 165 & 140 & 30 & 25 & 101 & 82.4 & 84.8 & 83.6 \\
Grok & GPT & 299 & 175 & 136 & 15 & 39 & 109 & 90.1 & 77.7 & 83.4 \\
Grok & Grok & 300 & 159 & 128 & 31 & 31 & 110 & 80.5 & 80.5 & 80.5 \\
Grok & Gemini & 296 & 164 & 142 & 22 & 22 & 110 & 86.6 & 86.6 & 86.6 \\
Gemini & GPT & 299 & 175 & 151 & 19 & 24 & 105 & 88.8 & 86.3 & 87.5 \\
Gemini & Grok & 300 & 159 & 136 & 37 & 23 & 104 & 78.6 & 85.5 & 81.9 \\
Gemini & Gemini & 297 & 165 & 151 & 25 & 14 & 107 & 85.8 & 91.5 & 88.6 \\
\bottomrule
\end{tabular*}
\caption{Full 300-record confusion counts and metrics from the raw judge run. Metrics are percentages. The main paper uses the 275 validated-record matrix.}
\label{tab:full-metrics}
\end{table*}

\FloatBarrier

\subsection{Sensitivity and Human Evaluation Findings}
\label{app:sensitivity}

These sensitivity checks ask whether the main paired conclusions depend on excluding diagnostic records or on refusal-heavy and conflict-aware generator behavior. The answer is no for the recall conclusion: the full-set estimate is almost identical to the validated-only estimate, and the behavior-sliced checks show that the near-zero global recall effect is not an artifact of refusals, conflict-aware answers, or unrelated outputs.

\begin{table}[H]
\centering
\scriptsize
\begin{tabular*}{\linewidth}{@{\extracolsep{\fill}}lccc@{}}
\toprule
Set & Recall $\Delta$ & \makecell{Avoided-claim\\flag $\Delta$} & Flag-rate $\Delta$ \\
\midrule
Validated (275) & -0.5 [-2.7,+1.7] & -4.3 [-6.6,-2.0] & -2.2 [-3.8,-0.6] \\
Full (300) & -0.7 [-2.8,+1.4] & -4.2 [-6.3,-2.1] & -2.2 [-3.8,-0.7] \\
Diagnostic (25) & -2.4 [-8.8,+2.6] & -2.9 [-6.9,+0.0] & -2.7 [-6.7,+0.7] \\
\bottomrule
\end{tabular*}
\caption{Global paired effects by dataset slice. The validated and full results agree; diagnostic records are small and uncertain.}
\label{tab:sensitivity}
\end{table}

\input{tables/behavior_paired_sensitivity}

\begin{table}[H]
\centering
\scriptsize
\begin{tabularx}{\linewidth}{@{}L{1.25cm}rY@{}}
\toprule
Automatic cell & Reviewed & Human finding \\
\midrule
TP & 23 & 21 genuine TP; 1 right verdict/wrong reason; 1 label mistake \\
FP & 25 & 22 alternate errors; 2 label mistakes; 1 unclear; 0 false alarms \\
FN & 9 & 6 genuine misses; 3 labels too broad \\
TN & 31 & 31 genuine TN; 0 missed errors \\
\bottomrule
\end{tabularx}
\caption{Expanded human-evaluation findings for the 88 reviewed cases.}
\label{tab:human-eval-findings}
\end{table}

\FloatBarrier

\subsection{Localization and Reproducibility}
\label{app:localization-table}

The final support check is whether judge flags localize to the expected evidence. Table~\ref{tab:localization} reports this in two ways: whether the cited source lies in the modified-passage set, and whether the free-text wrong-claim string explicitly contains the replacement value.

\begin{table}[H]
\centering
\scriptsize
\begin{tabularx}{\linewidth}{@{}L{1.0cm}Y Y@{}}
\toprule
Judge & Citation locality & Wrong claim contains replacement \\
\midrule
GPT & Gen=GPT 94.0; Gen=Grok 95.4; Gen=Gemini 95.0 & Gen=GPT 75.4; Gen=Grok 71.0; Gen=Gemini 83.6 \\
Grok & Gen=GPT 98.4; Gen=Grok 98.3; Gen=Gemini 98.5 & Gen=GPT 76.5; Gen=Grok 74.2; Gen=Gemini 82.4 \\
Gemini & Gen=GPT 95.3; Gen=Grok 95.5; Gen=Gemini 97.3 & Gen=GPT 76.8; Gen=Grok 72.8; Gen=Gemini 81.5 \\
\bottomrule
\end{tabularx}
\caption{Localization quality across matrix cells in the full-set analysis. Citation locality is consistently higher than exact replacement containment.}
\label{tab:localization}
\end{table}

\begin{sloppypar}
\raggedright
\footnotesize
\textbf{Reproducibility scripts.}
\path{audit/paired_audit.py};
\path{audit/behavior_stratified.py};
\path{behavior_paired_sensitivity.py};
\path{audit/build_cell_review_queue.py}.
Paired tables use seed 20260627 and 10,000 cluster-bootstrap replicates by core ID.
Checks verify 300 records, 275 validated records, 897 labeled outputs, and 2,683 judge verdicts.
\end{sloppypar}

%% file: tables/behavior_paired_sensitivity.tex
\begin{table}[H]
\centering
\scriptsize
\setlength{\tabcolsep}{2.5pt}
\begin{tabularx}{\linewidth}{@{}Y r c c c@{}}
\toprule
Slice & Answers & Recall $\Delta$ & \makecell{Avoided-claim\\flag $\Delta$} & \makecell{All-answer\\flag $\Delta$} \\
\midrule
All validated & 819 & -0.5 [-2.7,+1.7] & -4.3 [-6.6,-2.0] & -2.2 [-3.8,-0.6] \\
Context-follow only & 403 & +0.1 [-2.0,+2.3] & -- & +0.1 [-2.0,+2.3] \\
Context + both-claims & 448 & -0.7 [-2.9,+1.6] & -- & -0.7 [-2.9,+1.6] \\
No refusal, conflict, or unrelated & 462 & -0.7 [-2.9,+1.6] & +0.0 [+0.0,+0.0] & -0.6 [-2.8,+1.6] \\
\bottomrule
\end{tabularx}
\caption{Behavior-sliced paired sensitivities on validated records. Deltas are matching judge minus the mean of the two non-matching judges on the same answer, in percentage points with 95\% cluster-bootstrap CIs. Avoided-claim flag $\Delta$ is omitted when the slice has too few avoided-claim answers by construction.}
\label{tab:behavior-paired-sensitivity}
\end{table}